# Conducting Stylistic Analysis of Paintings through an Art-History Agent

Marc S. Walton,[1]* Astrid Harth,[2]*
[1] Museum Studies Programme, The University of Hong Kong, Hong Kong SAR
[2] Department of Chinese and History, City University Hong Kong, Hong Kong SAR
*Corresponding Authors

## ABSTRACT

Attributing an artwork to an artist has traditionally relied on detailed visual observations and descriptions, known as stylistic analysis in art history. By contrast, current artificial intelligence (AI) models used in the field offer only unexplained probabilistic classifications. To bridge this methodological gap, we present an AI framework that automates stylistic analysis of paintings, providing a foundation for enhancing evidence collection, discovery, and verification. By training a vision transformer (ViT) on a large corpus of paintings with metadata, our system encodes this art history-specific data as embeddings. These representations are factorized via sparse dictionary learning into a shared set of features that recur across the training set. A large language model (LLM) then interprets each feature by retrieving associated artworks and their accompanying curator-written texts, and synthesizes them into descriptions that reflect their stylistic attributes. Finally, an autonomous coordinator LLM applies a reasoning-and-action (ReAct) framework to weight, test, and refine these features into cohesive descriptions of an artwork, or comparisons of artworks. This approach converts detailed visual features into descriptive terms, addressing a key challenge in art history. It thus connects the use of images as data with the semantic concerns of humanists, establishing vision-based computational art history as an area for future growth.

## INTRODUCTION

From the late nineteenth century onward, art history increasingly adopted close visual analysis to determine attribution, stylistic periods and movements. Drawing on his medical training, Giovanni Morelli [1] treated the identification of an artist's hand as an anatomical classification task, focusing on seemingly minor details such as how ears were depicted in a painting as a marker for a given artist. Later scholars, including Bernard Berenson [2,3], Bruno Meyer [4], and Heinrich Wölfflin [5-7], extended visual analysis, aided by photographic reproduction, magnification, and the side-by-side projection of artworks, eventually establishing comparative analysis, using photographic reproductions, as a fundamental aspect of art historical practice and its starting place for writing about art and its history. Called stylistic analysis [1,8] of an artwork, these written descriptions capture the artist's use of formal, visual, and expressive features. Going beyond questions of just attribution, they also connect an object's visible and material characteristics to its history of production and to what Pierre Lemonnier [9] called an anthropology of technical systems linking objects, materials, practices, and social relations.

Somewhat surprisingly, the descriptive methods historians of art use to record and compare visual evidence [10] have changed relatively little since the discipline's formative period over 100 years ago. Thus, our study is motivated by the possibility of advancing stylistic analysis through AI. While AI vision models have achieved impressive accuracy in classifying paintings from digital images [11-19], their outputs cannot be traced back to the visual criteria that make an artwork an artwork, which has justifiably concerned art historians [20,21]. The main issue is that model classification probabilities generally lack calibration against visible evidence from the artwork itself [22,23].

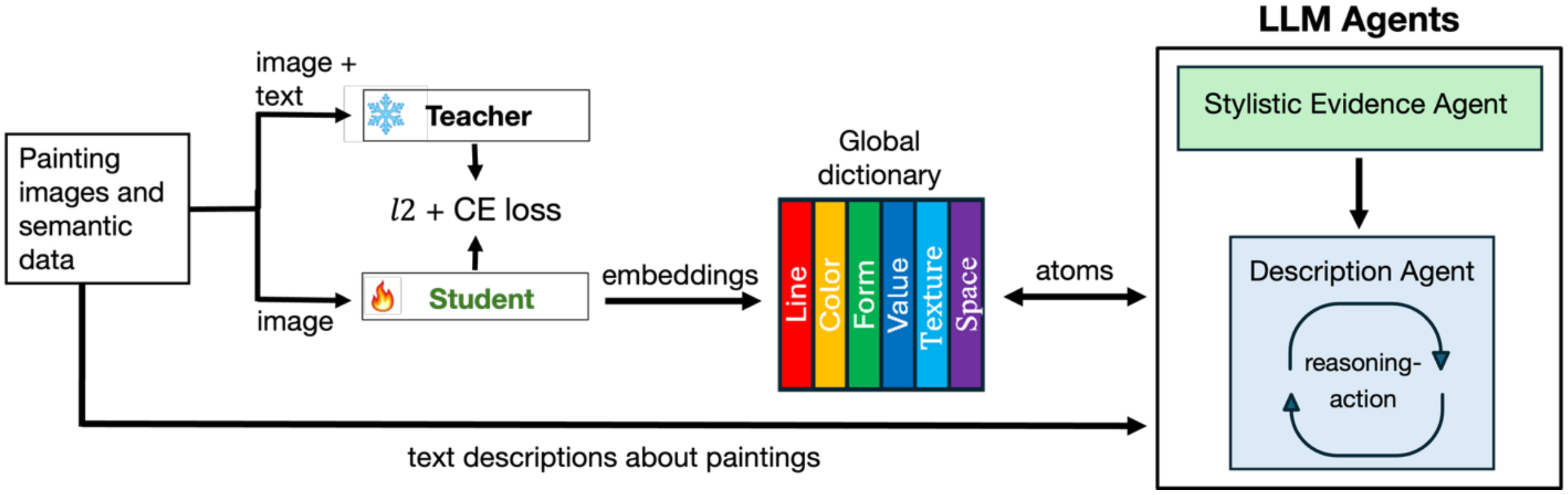

***Figure 1:*** Workflow for The Visual History Agent. The teacher concatenates images and text. The student learns the teacher's multimodal content and outputs ViT embeddings from images only. Sparse dictionary learning generates a set of visual atoms from the embeddings for the entire corpus of paintings. These data, together with curatorial written texts, are passed to LLM agents: A Stylistic Evidence Agent filters the associated semantic evidence through retrieval-augmented contextual summarization and synthesizes reusable labels. From these labels, the Description Agent produces extended stylistic descriptions for a given work of art.

We instead propose a specialized agentic system called the *Visual History Agent* (Fig. 1). This system combines a ViT with LLM agents to develop a vision-capable large language model for art history-specific tasks. Vision models are typically poor at consistently seeing and recognizing the context of objects [24]. Therefore, our main contribution is to demonstrate how to put together a framework capable of generating fully written stylistic analyses grounded in visual data. Put another way, we demonstrate an agent-based approach that translates visual information into words that make sense to art historians. Likewise, the broader implications of such an agentic system have general applicability in image-as-data disciplines such as microscopy, remote sensing, or even radiology.

The ViT "sees" artworks and enables comparison through their shared and differentiating visual characteristics. Dictionary learning factorizes the vector representations, or embeddings, produced by the ViT into a set of shared visual basis elements, or atoms. LLM agents then translate these atoms into human-understandable language by filtering and weighting pre-existing art historical texts associated with the relevant paintings. The two agents, which we call the *Stylistic Evidence Agent* and the *Description Agent*, are coordinated to prioritize evidence concerning visual form and content while stopping short of interpreting the artwork's broader contextual-historical meaning to conform to traditional stylistic analysis [25]. Rather than replacing the expertise of art historians, the agent should be regarded as a computational tool that makes observations and generates hypotheses for further downstream art-historical inquiry aimed at assigning artworks to time periods, cultures, artistic circles, workshops or individual artists.

## RESULTS

**Training Vision Model and Extracting Embeddings.** The first requirement for agentic stylistic analysis is a visual representation sensitive to the characteristic image contrasts, spatial features, and surface structures of paintings. Paintings occupy a restricted domain within the broader distribution of natural images [26], with compressed dynamic ranges and spatial-frequency characteristics [27-29]. To learn painting-specific regularities rather than relying solely on features acquired from natural-image datasets such as ImageNet, we fine-tuned a ViT on a corpus of paintings spanning most periods, regions, and workshop traditions (Fig. 1 and Materials and Methods).

Each training image was associated with structured metadata (e.g., artist, date, and period) and descriptive text concerning its subject matter and stylistic properties. We refer collectively to these data as the painting's semantic information. Through a process called semantic anchoring, we concatenate an image embedding with text embeddings to produce supervisory targets within a teacher–student distillation framework [30].

The resulting multimodal teacher representation relates paintings to their art-historical descriptions, while the student learns to reconstruct that representation from image input alone. At inference, the student therefore functions as an image-only encoder that produces an embedding combining learned visual and semantic characteristics. These embeddings support structured interpretation by language models once the dictionary-atom interpretation pipeline connects these latent dimensions to natural-language descriptions.

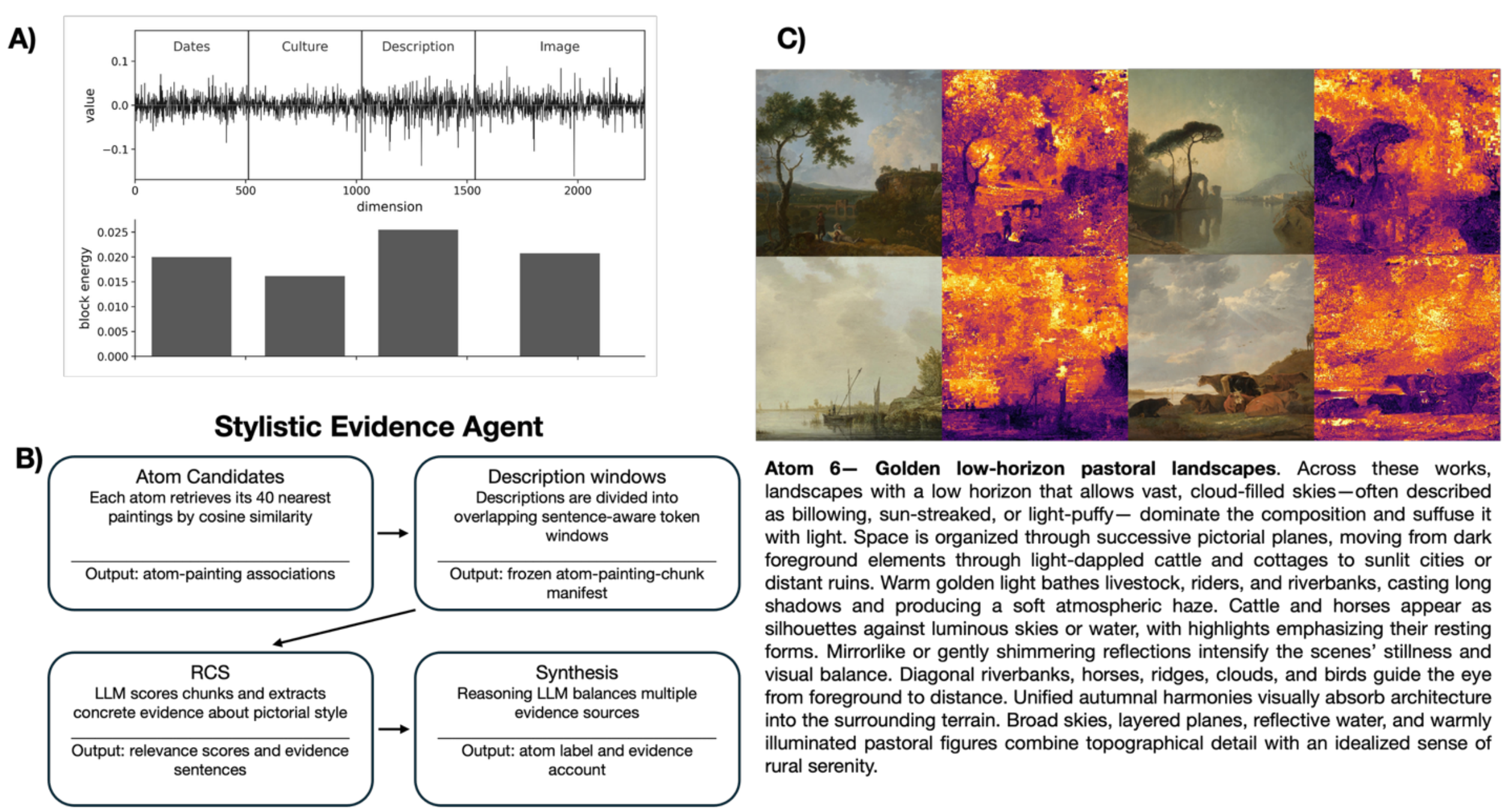

***Figure 2:*** Stylistic Evidence Agent: A) An atom and its corresponding semantic block energies, whose structure results from training. B) The top candidate descriptions per atom are aggregated using retrieval-augmented contextual summarization (RCS) and then reranked by a reasoning model before the atom label is written. C) Atom 6 is dominated by characteristic scenes featuring low horizons, billowy clouds, and landscapes. The coefficient weights indicate where this concept is most expressed on the picture plane in the top four paintings activated by this atom. The top two paintings are, respectively, by Richard Wilson (1762), a Welsh painter, and Albert Cuyp (1640), a Dutch painter. The bottom two paintings are by Wilson (18th century) and Cuyp (1645). Although these paintings date from different time periods and cultural regions, Wilson was known to collect Cuyp's work, so Cuyp likely influenced the Welsh painter's work.

**Learning a Corpus-Wide Dictionary.** A key insight is that no single visual feature or contrast is unique to any given painting. Instead, visual patterns tend to recur across artworks regardless of historical period or cultural attribution, suggesting that paintings can be characterized by shared directions within the embedding space. To expose these shared directions, we collectively decompose all embeddings across the corpus into linear combinations of atoms [31]. We perform this step with K-SVD, an iterative dictionary-learning method [32-35], which summarizes the dataset's common features and contrasts into a dictionary (Materials and Methods).

Among the most consequential design parameters for dictionary learning is encouraging sparsity, which enhances the isolation of distinct features. Non-negativity further constrains the representation to additive combinations, supporting interpretation. Based on this, we parametrized the global dictionary with 1,000 atoms, which provides the best empirical balance for reconstruction accuracy, sparsity, atom stability, and interpretability. We then encode each painting against the fixed dictionary [36-38]. The resulting coefficients identify the atoms contributing to the reconstruction of an embedding from a given painting, with coefficient magnitude indicating their relative contribution. Each painting is therefore approximated as a sparse, non-negative combination of visual features learned across the corpus.

As exemplified by Atom 6 in Fig. 2a, its 2304-dimensional embedding space preserves the block structure introduced during ViT training, so each atom can be partitioned into components corresponding to the visual representation and the three semantic-information groups, including date, culture, and description (Materials and Methods). This enables block-wise analysis of the atom. Atom 6, for instance, has its greatest vector energy in the description block, indicating that the associated texts were particularly influential in determining the atom's learned structure. This example demonstrates how training with semantic anchors shapes the representational structure learned by the ViT and helps connect visual features to art-historical language.

The meaning of an atom is conveyed from a numerical sequence into natural language by identifying the textual evidence most strongly linked to it. This is accomplished through the *Stylistic Evidence Agent* (Fig. 2b). The agent first uses cosine similarity to retrieve the paintings whose embeddings are most closely aligned with a given atom. The descriptive texts associated with the top paintings per atom are then divided into overlapping textual chunks and processed by a compact local LLM (Qwen2.5-7b) using retrieval-augmented contextual summarization (RCS) [39]. The modest model size permits local processing of the thousands of chunks associated with the whole dictionary. The LLM scores each chunk for relevance, prioritizing specific observations concerning line, contour, color, illumination, modeling, depth, texture, surface, drapery, and composition. Chunks concerned primarily with provenance, dating, iconography, or narrative receive lower scores.

An online reasoning LLM (GPT-5.1) then evaluates the candidate chunks collectively, considering both their relevance scores and the recurrence of similar language terms and patterns. The model then re-ranks the evidence to best align with extracting stylistic concepts before generating a concise label and a description of the features that form an atom. A claim for a stylistic attribute is retained only when supported across multiple paintings, prioritizing recurrent evidence over isolated details or terminology. For Atom 6, the most prominent features mentioned are the sky, clouds, and low horizon plane, as per the associated text in Fig. 2C. These features are not common to a particular time period, as indicated by the top four activated images being from the 17$^{th}$ and 18$^{th}$ centuries and belonging to two different geographic regions (Wales and the Netherlands), though this could indicate the cultural transfer of style [40].

As the ViT preserves a spatial field of local patch embeddings, the global dictionary can also be mapped to the picture plane. Each patch is encoded against the complete dictionary, producing a coefficient map for every atom as exemplified by the heat maps in Fig. 2C (Materials and Methods). These maps show weaker coefficients in figural content, architectural features, and specific elements like trees that act as subjects rather than the surrounding broader landscape, thus aligning with the associated text.

The *Stylistic Evidence Agent* therefore serves complementary purposes: (1) the coefficients from the sparse encoding quantify how strongly each shared atom contributes to an embedding, while (2) the retrieved textual evidence provides a testable interpretation of what that atom may represent. The embedding-to-atom, atom-to-chunk, and atom-to-label associations are frozen as reusable corpus-wide references, ensuring that subsequent paintings are analyzed against the same dictionary and descriptive evidence. This inspectable vocabulary of shared features provides the substrate for the painting-level and comparative analyses that follow.

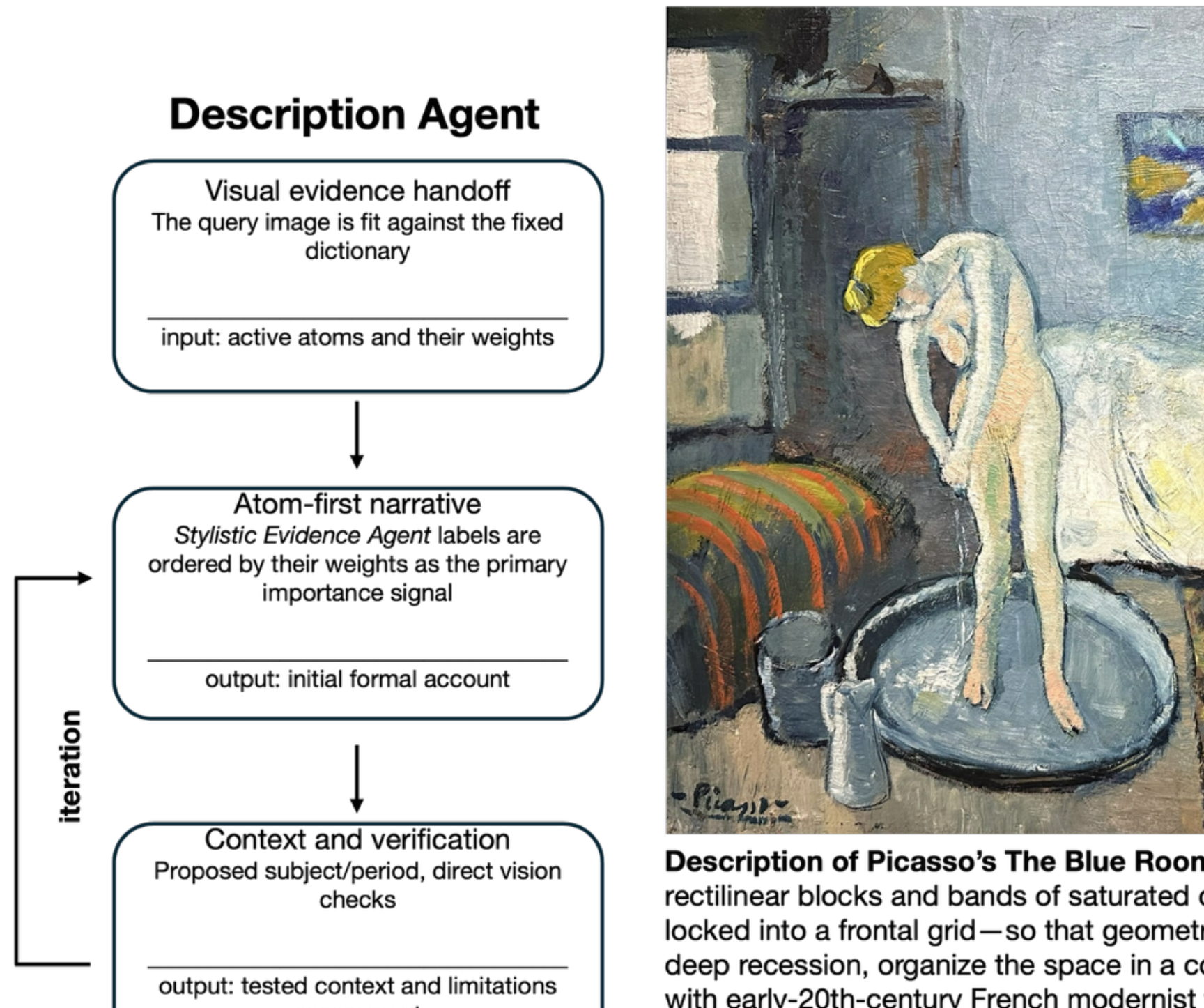

***Figure 3:*** Description Agent. ReAct framework that combines atomic evidence to describe a painting's composition. The pipeline takes us from the atoms associated with Picasso's Blue Room (1901, Phillips Collection) to a fully contextualized description. In the description generated by the Description Agent, sentences are supported by the top-weighted atoms, shown in bold. The nearest neighbor provides the French modernist context. Vision, from GPT 5.1, supports the scene structure and analyzes the painting's condition. However, the primary evidence for the painting's contents comes from the atoms.

## DISCUSSION

**Writing the Description.** Once the global dictionary and its semantic labels have been established, these data may be applied to a painting *not included* in training, as demonstrated using Pablo Picasso's *The Blue Room*, 1901, from the Phillips Collection (Fig. 3). Picasso provides a challenging test because his artistic practice encompasses markedly different formal and stylistic modes throughout his career - ranging from realism and cubism to surrealism. The image alone is first processed by the ViT to produce an embedding that is encoded against the fixed dictionary [41].

The active atoms, along with their previously generated labels and descriptions from the *Stylistic Evidence Agent*, are then passed to our *Description Agent*, a reasoning model (coordinated by GPT-5.1). The atom and their associated coefficient weights are treated as the primary measure of evidential importance. The *Description Agent* first re-ranks the atoms according to their usefulness for interpreting the painting, retaining the complete active set while prioritizing atoms with larger coefficients. It then synthesizes an initial stylistic description. At this stage, the model receives neither metadata nor the description of stylistically related works. The initial description is therefore grounded in the shared atom vocabulary rather than in a prior assumption about the painting's subject, date or authorship.

Only after this preliminary atom-based account is established is the date and cultural context introduced. A nearest-neighbor search over the embeddings retrieves the closest five paintings in the reference corpus. The coordinating agent considers only their structured metadata (artist, date, and period) but never accesses the nearest neighbor's descriptive texts. These fields provide provisional attribution hypotheses. For Picasso's *The Blue Room*, the nearest neighbor was André Lhote's Cubist-style study for *Homage to Watteau*, dated approximately 1918. Although this date is 17 years after Picasso's 1901 painting, Lhote was influenced by Picasso and was a member of his artistic circle, so his work inherits some of the earlier painting's stylistic qualities [42]. This result shows that nearest-neighbor localization is valuable for art historical classification and can be extended to establish broader artistic connections.

The vision-capable reasoning model also examines the query image to directly establish basic visible facts, including the number and relationships of figures, prominent objects, compositional setting, and observable surface conditions, such as brushstroke characteristics or craquelure. These observations test the visual compatibility of neighbor-derived hypotheses and help interpret the atom evidence. As the *Description Agent* describes *The Blue Room*,

> *A single nude bather stands in a round basin near the center-left, functioning as the dominant central motif on a relatively open ground of floorboards and bedcover, so that broad, emptier planes frame the body while the shallow, almost relief-like interior space seems to pin her to the picture plane*

With a subject of a nude bather, the direct-vision evidence here has conditioned the atom evidence by requiring a specific figure, object, or setting to be present in the final write-up. Explicit guardrails, however, prevent this vision stage from hallucinating [24], which would otherwise displace the atom-derived stylistic description with an unrestricted description of the image.

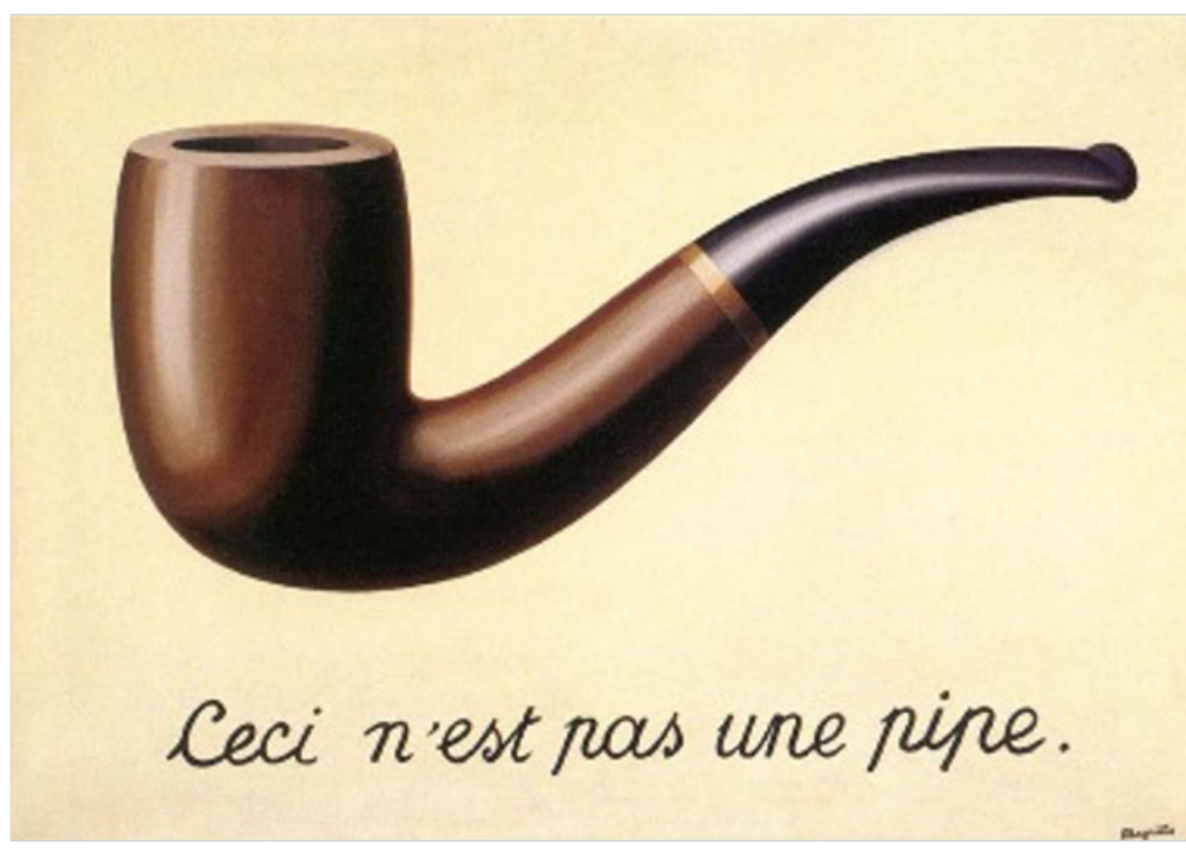

A single smoking pipe is centered against a broad, light, warm-toned ground, its smoothly modeled bowl and stem reading as a small, tactile volume held away from an otherwise featureless field [**Atoms: 1, 240, 288**] [**Vision**]. The pipe's contour is crisply drawn yet gently shaded, so that the darker stem and warmer wooden bowl emerge from soft tonal gradients rather than sharp local color contrasts, giving the object a quietly atmospheric presence akin to heads or still-life motifs that are made to stand out through controlled chiaroscuro rather than saturated hue [**Atoms: 1, 161, 193, 240**]. Below this emblematic form, a single horizontal line of cursive text stretches across the surface like a ruled band on a manuscript page, turning the lower register into a frieze of handwriting that balances the isolated image above while recalling the disciplined, line-by-line structure of manuscript script fields [**Atoms: 72, 108, 146, 956**] [**Vision**]. The relationship of object and script recalls a highly reduced still life, in which a lone, carefully lit utensil—akin in focus to reflective vessels or quiet banquetware—is suspended on a shallow stage without cast shadow, its polished curves suggested more by tonal sweep than by descriptive detail [**Atoms: 1, 193, 240, 288**]. The background functions as a flat, continuous plane, closer to a muted color field than to deep space, so that the pipe seems to float low within a dominant, luminous expanse much as a small motif might read against a wide sky band above a low horizon in horizontally organized vistas [**Atoms: 21, 131, 240, 550**]. Line itself becomes a primary event along the lower edge, where looping, rhythmic script introduces a band of animated strokes that contrasts with the pipe's steadier outline yet remains anchored to the same shallow, poster-like plane, echoing patterns in which linear ornament or tracery plays across flattened grounds [**Atoms: 72, 108, 956**]. The painting can be understood as a conceptual still life that stages a dialogue between a familiar object and its written designation, making the tension between image and text the central focus rather than any broader narrative subject [**Atoms: 21, 72, 146, 193**] [**L2**]. Around the pipe's silhouette, the softly feathered transitions of tone suggest a light, almost satin-like envelope of air, with delicate, fluttering shifts in value that recall feathery handling and shimmering highlights used to suggest rustling textiles and airy landscapes [**Atoms: 152, 161, 240**]. In a way that can plausibly be related to mid- to late-20th-century non-narrative painting, the restrained palette, planar background, and single centrally aligned motif resonate with modern experiments in rectilinear formats and minimal surface incident, even as the softly graded modeling and atmospheric transitions keep the work from becoming entirely hard-edged or purely abstract [**Atoms: 21, 240**] [**L2**]

***Figure 4:*** The Treachery of Images by René Magritte (1929) and the associated output from the Description Agent.

The evidence is integrated through iterative reasoning-and-action cycles (Fig. 3; see Supplementary Data 1 for complete text description, prompts, and reasoning-action evidence). During each cycle, the model revises the description according to a fixed evidence hierarchy in which coefficient-weighted atoms remain the primary evidence, and the previously tested visual and metadata evidence serves to constrain and support. A subsequent observation stage audits the text for unsupported claims, inaccurate atom citations, excessive reliance on direct vision, omitted high-weight atoms, and the reintroduction of rejected neighbor-derived details.

At least 75% of sentences must cite atom evidence, no more than 25% may cite direct vision, and the high-weight atoms collectively accounting for approximately 85% of coefficient mass must receive substantive treatment. These requirements, together with sentence-level citation validity, are checked programmatically, and revision continues until no unsupported or overstated claims remain. Every sentence in the final account therefore identifies its supporting atoms, direct observations, or contextual hypotheses. These constraints are illustrated by another example from René Magritte's 1929 painting *The Treachery of Images* located at the Los Angeles County Museum of Art. Here, the *Visual History Agent* produces an approximation of a human-generated formal description (Fig 4). Not only does the description closely match the painting's visible content, but it also conveys deeper meaning, particularly through the text "ceci n'est pas une pipe" that sits below an image of a pipe. The agent writes that this line of text itself becomes "a primary event" of the image, indicating a capability to juxtapose, compare, and assess the function of this written content [22].

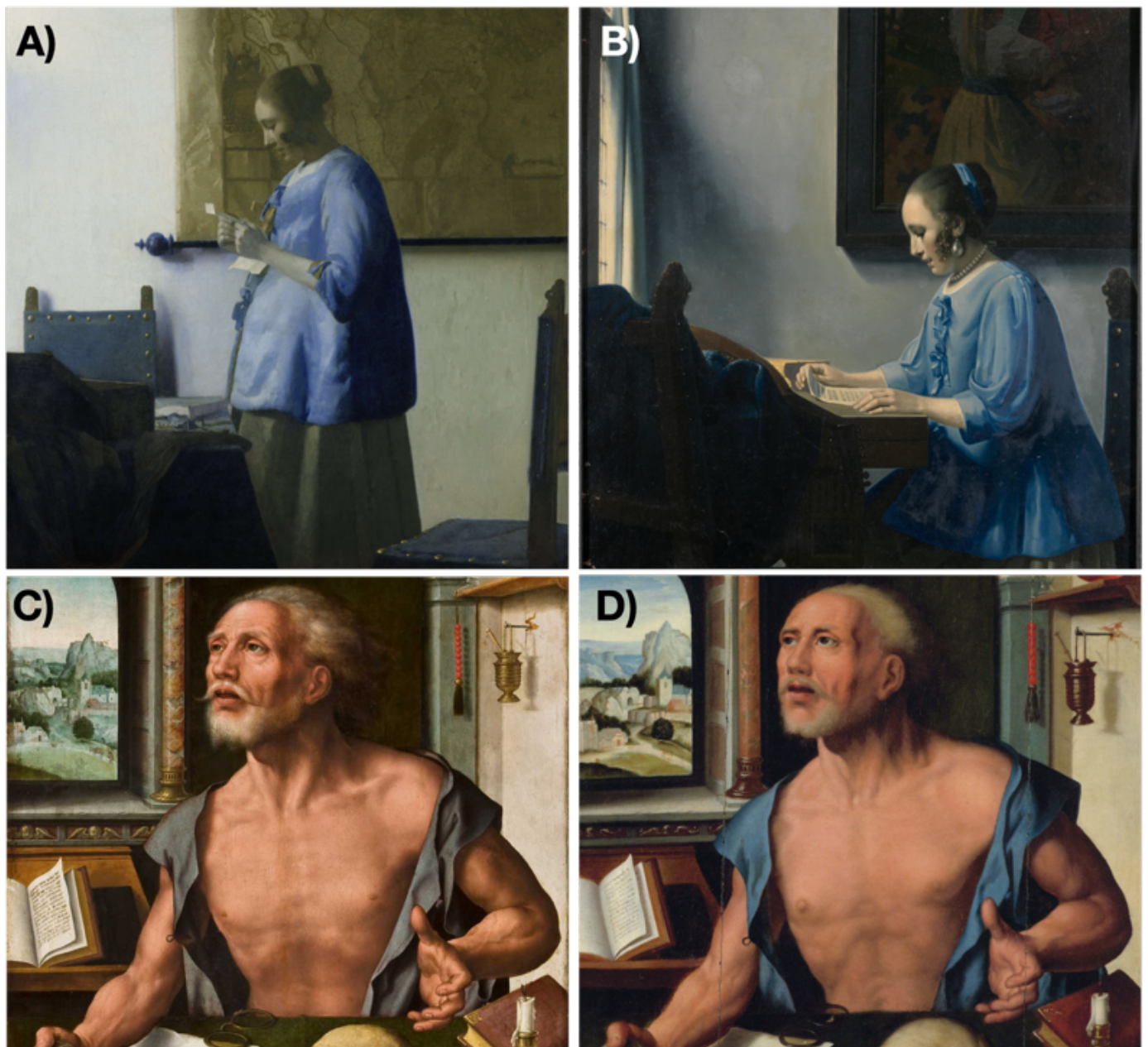

***Figure 5:*** Close comparison and contrast capabilities. A) and B) show, respectively, a Vermeer and a van Meegeren fake Vermeer (Rijksmuseum) that may have served as inspiration for the former. C) shows a workshop piece from Joos Van Cleve from the 16th century and D) a later copy probably dating from the 17th century. The corresponding comparative texts generated by the Description Agent can be found to the right of the images.

A central application of this framework is indeed the comparison of paintings that share a subject, composition, or a proposed attribution to a given artist (Fig. 5 and Supplementary Data 2 and 3). This application makes the agentic framework particularly relevant for examining copies, workshop productions, pastiches, as well as forgeries. The ViT processes each image independently and encodes it against the same fixed dictionary, producing non-negative coefficient vectors within a common coordinate system. Atoms active in both paintings identify shared representational tendencies, whereas atoms active in only one indicate potential stylistic differences. Differences in normalized coefficient mass further reveal whether a shared atom is dominant in one work but contributes only weakly to the other.

Using this approach, we examine what is today a trivial test case but reflects the state of art-historical knowledge in the 1930s: Vermeer (Fig. 5A) and a known forgery of Vermeer by Han van Meegeren (Fig. 5B) [43,44]. The text produced by the *Visual History Agent* shows that it is capable of placing the forgery in a modern context and successfully contrasts this with the actual seventeenth-century interior genre scene by the Dutch artist Johannes Vermeer. In another more subtle example, looking at a sixteenth-century Netherlandish painting attributed to the workshop of Joos Van Cleve (Fig. 5C), the model contrasts this to a similar work in the collection of the Princeton University Art Museum (Fig. 5D). As has been discussed by Rapoport [45], the relationship between these artworks is that the latter is likely an executed work by a later copier dating from the seventeenth century. The *Visual History Agent* likewise suggests a Baroque date for the Princeton artwork and notes simplified anatomical and object detail compared to the earlier sixteenth-century workshop painting, which retains finer details. These comparisons suggest that the *Visual History Agent* can first make a painting's stylistic qualities explicit and do so in a manner that approaches human-level expert judgment.

The *Visual History Agent* framework more concretely situates computational art history within the broader digital humanities by treating stylistic properties of artworks as directly inspectable evidence. Rather than reducing a painting to a single attribution score, as inherent to a classification problem, we separate analysis into examinable stages, allowing art historians to follow how the system moves from image features to stylistic interpretation through comparative analysis.

We view this introduction of vision-based agentic AI into art history as one with a structural parallel in the twentieth-century adoption of scientific imaging and analytical techniques that contributed to the development of technical art history. X-radiography, introduced to painting analysis in the early twentieth century [46], and infrared reflectography, developed more fully during the 1970s [47], expanded the range of evidence available to art historians by revealing aspects of materials, construction and artistic process inaccessible to ordinary vision. This study extends this trajectory in a different direction by creating computational methods for automated stylistic comparisons. Art historians can use these methods to verify stylistic evidence, examine large digital image collections, identify common visual features, and systematically compare patterns, all while maintaining transparency about the evidence behind each computational interpretation.

More broadly, our research also improves how we treat images as data. With art history's extensive written records, semantic labels assigned to paintings significantly improve ImageNet-based embeddings, making them more relevant and specific to art history as a subdomain. Furthermore, using dictionary learning to analyze similarities and differences in embeddings across large datasets yields interpretable insights in natural language terms, a method adaptable and relevant to many other scientific disciplines.

## METHODS

**Painting Corpus and Metadata.** This study was intentionally restricted to paintings, resulting in a database of 39,353 digital images. These images, together with their metadata, were compiled from museums that provide collection information and digital images through open-access programs, public application programming interfaces (APIs), bulk-data releases, or linked-open-data services. The principal institutional sources included those from The Metropolitan Museum of Art, the Rijksmuseum, the National Gallery of Art, Washington, DC, the Cleveland Museum of Art, the Smithsonian Institution, the National Gallery, London, the Musée du Louvre, the J. Paul Getty Museum, Tate, and the Art Institute of Chicago. Smithsonian records comprised the National Museum of Asian Art, National Portrait Gallery, Hirshhorn Museum and Sculpture Garden, and National Museum of African Art. While there may be additional institutions that could have been included, this body of images spans a wide range of dates from antiquity to the present day and encompasses cultures, enough to be considered encyclopedic in scope.

Each record retains the institution’s canonical collection information for a given artwork, allowing the locally stored image and database content to be traced back to its source record. The unified metadata schema contains the artwork title, attributed artist, date, art-historical period, medium, culture, region, holding collection, source-record URL, and descriptive text. Extended curatorial catalogue entries that are relied on to produce stylistic descriptions are present for 46.3% of the corpus or 18,216 paintings. When fields of metadata were missing for an artwork, they remained as absent values rather than being statistically imputed, and original institutional terminology and source links were retained where possible. A complete repository of this information is provided as a Microsoft Excel file in the project's GitHub repository: https://github.com/homelyprotestant/Visual_History_Agent.

**Vision transformer and semantically enriched image embeddings.** The vision model used in this study was trained with a teacher–student semantic-distillation framework that we previously described for optical microscopy [30]. As shown in the UMAP visualization in Supplementary Fig. 1a,b, this framework anchors image embeddings to textual descriptions of the artworks, biasing the embedding geometry toward art-historically salient axes of variation rather than incidental correlates inherited from ImageNet-style natural-image pretraining. The resulting representation exhibits fine-grained structure that encodes culture, artist, chronology, and related attributes. Used as the student's training target, it therefore compels the model to learn contrasts that are useful for the history of art.

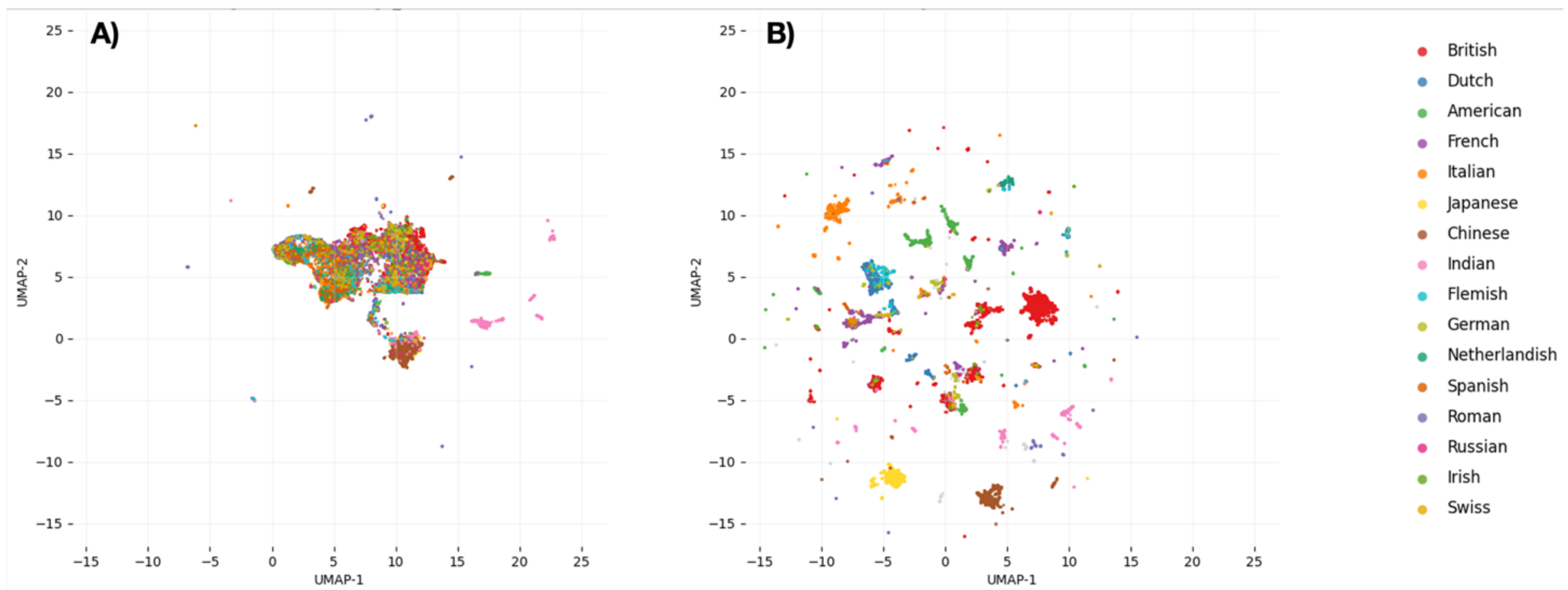

***Supplementary Figure 1:*** Teacher embeddings clustering as visualized by UMAP. A) Output without semantic anchoring. B) After semantic anchoring with metadata. Points in both plots are grouped according to the culture in which the painting was made.

As in our previous work, we used a pretrained LongCLIP ViT-B/16 backbone that has both image and text encoders [48]. Using the present model, we employed both encoders to enhance interpretation of the visual regularities in painted artworks. Specifically, the 768-dimensional visual embedding was extracted from the final ViT transformer block after the final layer normalization (ln_post) but before projection into the model's 512-dimensional image–text embedding space. For downstream atom analysis, we retained the corresponding 768-dimensional patch-token representations rather than relying on the CLS token. We normalized these patch tokens and mean-pooled them to form the visual block, while preserving the local representations needed to spatially interpret detailed painting features.

Semantic fields were encoded independently with the text encoder and organized into three 512-dimensional semantic blocks: (1) dates and art-historical period; (2) culture, region, and artist; and (3) painting title and descriptive text. The complete teacher embedding therefore contained 2,304 dimensions: the 768-dimensional visual block and the three 512-dimensional textual blocks concatenated onto each other. Encoding metadata fields independently prevented long descriptive passages from overwhelming shorter but informative fields. The block structure also allowed for further interpretation of where the available art-historical information came from in the pipeline sequence, allowing for better traceability and auditing.

From this information-rich teacher embedding, an image-only student was trained to reproduce the teacher's structure. The student, using the same ViT structure as the teacher, passed its 768-dimensional visual representation through a multilayer perceptron (MLP) with successive dimensions of 1024, 1536, and 2304. The final dimension matched the size of the teacher representation. Thus, the function of the MLP is to expand the student embedding into the multimodal teacher space.

Training minimized a weighted combination of an $l1$ reconstruction loss between the student and teacher embeddings and a cross-entropy loss over weak curriculum labels derived from the corpus metadata:

$$\mathcal{L} = \left\| z_s - z_y \right\|_1 + 0.11\mathcal{L}_{CE} \quad (1)$$

where $z_s$ and $z_y$ denote the student and teacher representations, respectively. The reconstruction term transfers the coordinate-level structure of the multimodal teacher space, whereas the classification term discourages representational collapse and maintains separation among historically and visually related groups.

Student training proceeded through three curriculum training passes based on increasingly refined weak-label partitions derived from painting metadata. These passes comprised 165, 467, and 683 pseudo-classes, respectively. The pseudo-classes were generated from a hierarchical metadata tree organized successively by culture, art-historical period, and artist. Branches were recursively subdivided, and leaves containing more than ten paintings were retained as pseudo-classes in the final pass, producing the 683 output logits. Samples belonging to unsupported leaves were assigned a noise label and excluded from the cross-entropy loss term in equation 1. Model weights were transferred between curriculum passes, while the classification layer was remapped to the labels defined for each new pass. The model was trained via progressive unfreezing, advancing from the visual readout, the final two transformer blocks, the final four blocks, the final eight blocks, all 12 blocks, and finally the complete model. Training was conducted in a CUDA-enabled Google Colab environment.

The checkpoint used in this study was obtained during the first epoch of the complete model training and achieved 86% validation accuracy across 2,987 paintings. This accuracy measures prediction of the weak curriculum labels rather than artwork attribution accuracy. For this reason, we considered a group of Italian Renaissance artists associated with Raphael who painted in the sixteenth-century (Perugino, Romano, Polidoro da Caravaggio, Parmigianino, Sebastiano Del Piombo, Sodoma, and a contrast set of nineteenth-century copies) that would usually require an expert’s knowledge to distinguish between. In Supplementary Fig. 2, we show the output of our ViT as a confusion matrix based on the final CE layer of the network, which correctly identified the attributions for 724 of the 730 paintings from the presented Renaissance cohort, achieving 98.9% accuracy across 9 classes, indicating that the semantically distilled representation retained fine-grained distinctions among Raphael and closely associated artists and copies.

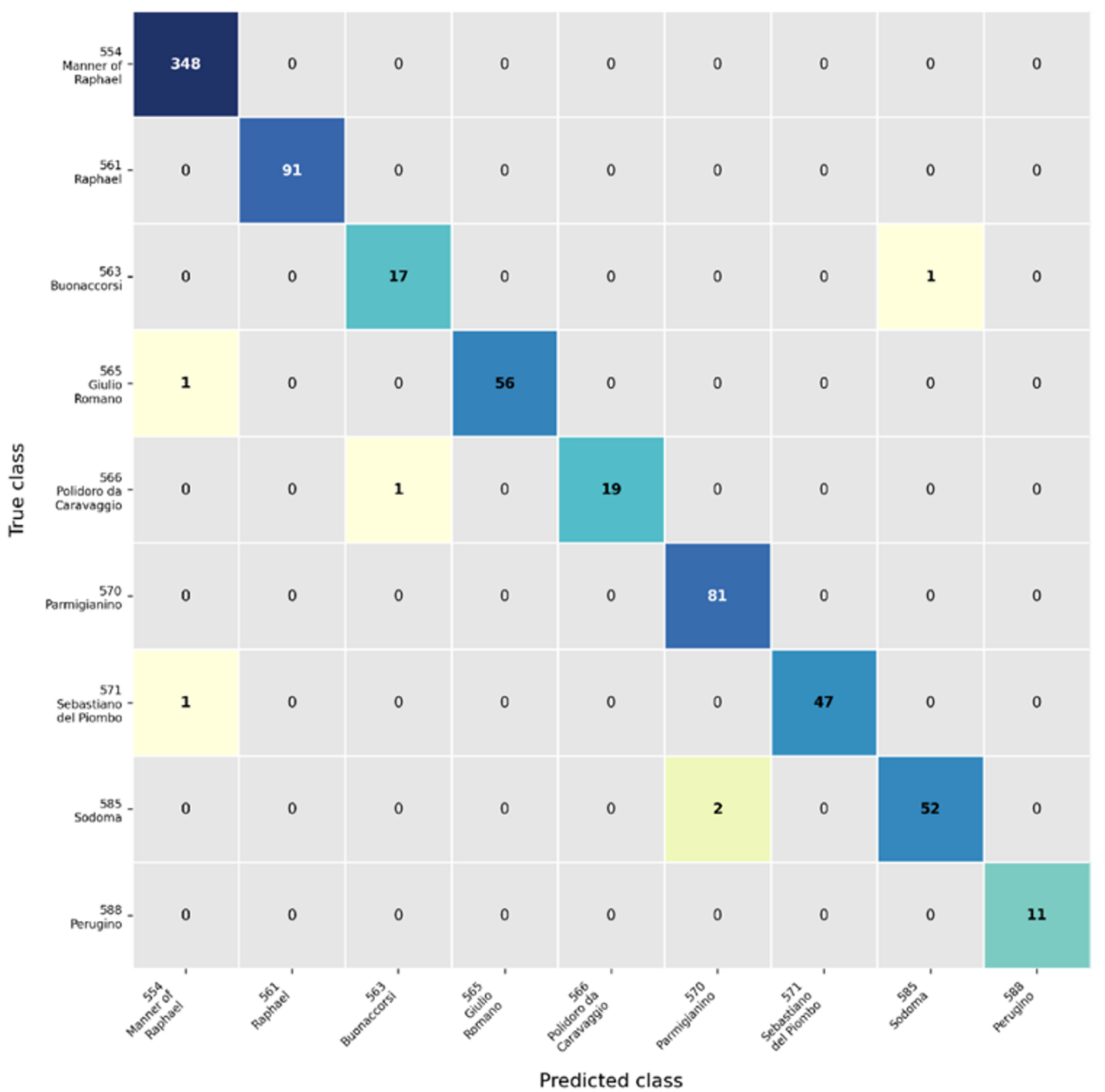

***Supplementary Figure 2:*** A confusion matrix demonstrating the image retrieval capabilities of the ViT model trained on an encyclopedic breadth of paintings. The test for retrieval performance was done a narrow group of Italian Renaissance Artists surrounding Raphael.

**Image Pre-crop and Inference**. Each painting was represented by a high-resolution central crop designed to preserve fine-scale cues relevant to stylistic analysis. Conventional downsampling of artworks to 224×224 pixels discards art-historically important information.[11] Images were therefore resized so that relatively high-resolution features are retained. This is accomplished by downsampling the shorter side to 1120 pixels, after which a centered 1120×1120 pixel crop was extracted. This crop was passed through the vision transformer to yield a 70×70 grid of local patch representations, each 16×16 pixels. To accommodate the larger spatial grid, the learned positional embeddings were interpolated from the model's native 14×14 layout to 70×70.

The relatively large input size balanced computational practicality with the spatial resolution required for art-historical close examination. The final transformer layer produced 4,900 patch embeddings, each initialized from the 16×16-pixel patches. Because of self-attention, each patch representation also integrates information from the wider image, so that local structure is contextualized by the painting as a whole. The local patch embeddings were $l2$-normalized and mean-pooled to produce a 768-dimensional embedding, followed by an MLP projection to the final 2304-dimensional size used in this study. These pooled embeddings were computed for the complete corpus and used for both FAISS-based image retrieval and subsequent dictionary learning.

The unpooled 70×70 patch field was retained for spatial visualization. Dictionary atoms were evaluated across this field, producing maps that identify the image regions in which each atom is most strongly expressed. These maps were super-resolved from 70×70 pixels to 224x224 pixels for display using a joint dictionary learning previously described [49].

**Approximate K-SVD learning of a dictionary.** The corpus-wide dictionary was learned from the mean-centered embedding matrix, $X \in \mathbb{R}^{2304\times 39,353}$, where each of the 39,353 columns corresponds to one painting and each row represents the 2304-dimensional embedding space. Mean centering was performed by subtracting the corpus-wide mean from every painting representation before dictionary learning. We sought a factorization of $X$ following,

$$X \approx WD \qquad (2)$$

where $D \in \mathbb{R}^{1000 \, x \, 2304}$ is a dictionary of unit norms and $W \in \ \mathbb{R}_{\geq 0}^{Nx1000}$ contains non-negative sparse coefficients. Non-negative sparse encoding was achieved with elastic net optimization[50] as expressed by

$$\min_{W\geq 0} \frac{1}{2}\|X - WD\|_F^2 + \alpha\left[\rho\|W\|_1 + \frac{1-\rho}{2}\|W\|_F^2\right], \quad \|d_k\|_2 = 1 \qquad (3)$$

where $d_k$ is the $k-th$ dictionary atom, and $\alpha$ is the elastic net strength and the $l1$ ratio $\rho$. No fixed number of active atoms was imposed.

The hyperparameters for learning the dictionary were chosen empirically by jointly scanning dictionary size and nonnegative Elastic Net strength, $\alpha$. Held-out reconstruction MSE showed the best reconstructions with $\alpha = \ 3x10^{-5}$ and $l1$ ratio $\rho = 0.5$. We found that a dictionary size of 1000 was optimal, with diminishing returns in reconstruction error at larger sizes.

Learning alternated between sparse encoding and dictionary updating steps. At each iteration, 10% of the corpus was sampled and encoded against the current dictionary using non-negative elastic net. Rather than performing the full singular-value decomposition used in exact K-SVD, the atom was updated by projecting a restricted residual onto its active coefficient vector:

$$d_k \leftarrow \frac{w_k^T E_k}{\|w_k^T E_k\|_2} \qquad (4)$$

where $w_k$ are the coefficient weights of atom $k$ across paintings in which that atom is active, $E_k$is the reconstruction residual after removing every atom except $k$, and the $w_k^T E_k$ term is the coefficient-weighted average direction of the residuals associated with atom $k$. The corresponding coefficients were updated by projecting the residual onto the revised atom. This approximate rank-one update substantially reduces the computational cost of learning from the full corpus while preserving the alternating sparse-coding structure of K-SVD.[51]

Atoms receiving no assignments were reinitialized from embeddings with large reconstruction residuals. Every three iterations, atoms supported by fewer than 20 sampled paintings or for which the five strongest activations accounted for more than 40% of their total coefficient mass were replaced using high-residual examples. We replaced at most 50 atoms in one pruning step. We renormalized dictionary rows after every update. Finally, we re-encoded all corpus embeddings against the fixed dictionary using equation 2.

The resulting dictionary atoms and painting-level sparse coefficient vectors were stored for downstream processing by the LLM agents. These stored representations bridge the framework's computer-vision and natural-language reasoning components.

**Corpus-level atom interpretation.** The meaning of each dictionary atom was interpreted by the *Stylistic Evidence Agent.* The paintings most strongly aligned with each atom direction in the mean-centered corpus embedding space were identified by computing cosine similarity between the unit-normalized atom and every unit-normalized painting embedding. For each atom, the title and/or description of the $K$ nearest paintings were retained as evidence for interpreting the atom's recurring stylistic meaning. We set $K = 40$ as a practical compromise that considered the balance between evidential breadth, computational cost, and the amount of material supplied to the language models.

The text associated with each retrieved painting was divided into sentence-aligned windows of approximately 180 tokens, with a 60-token overlap. Each unique window was then processed through RCS [39] using Alibaba's Qwen 2.5-7B, a compact local model selected for its ability to score tens of thousands of textual windows efficiently.

The model was prompted to assign relevance scores according to the window's usefulness for interpreting an atom's visual qualities. Scores of 8–10 indicated specific and detailed discussion of form, shape, line, contour, color, light, modeling, depth, texture, surface, drapery, composition, pattern, brushwork, layering, or finish. Scores of 4–7 indicated useful compositional or spatial description with limited formal detail. Scores of 1–3 indicated passages primarily concerned with provenance, attribution, dating, iconography, or narrative, unless such information clarified a visual feature.

An online coordination model (GPT-5.1) then adjudicated the material associated with the retrieved paintings for each atom. Candidate passages were reranked by jointly considering their cosine association with the atom, their specificity for stylistic analysis, and the recurrence of comparable observations across paintings. Cosine proximity and the local-model relevance score were treated as measures of atom association and textual usefulness, respectively, rather than as visual evidence in themselves. The coordinator subsequently generated a concise atom label and a four- to eight-sentence description of the recurring formal qualities associated with the atom.

The atom-to-painting retrievals, atom-to-text-window associations, local contextual summaries, coordinator rerankings, labels, and extended descriptions were stored as reusable corpus-level caches. Together with the dictionary atoms and painting-level sparse coefficient vectors, these records form the interface between the vision model and the downstream language-model reasoning system. The atom coefficients indicate which atoms contribute to a painting as well as the strength of those contributions, while the stored corpus-level interpretations provide natural-language explanations of the formal qualities represented by those atoms.

**Description Agent.** For a given painting query, the *Description Agent* encodes the image using the same student ViT process employed for the corpus and derives a non-negative elastic net sparse code relative to the global dictionary. The active atoms and their normalized coefficients represent the stylistic evidence associated with the painting. The labels for these atoms are retrieved from the fixed cache created by the *Stylistic Evidence Agent*, and reordered for synthesis, prioritizing coefficient magnitude as the key importance signal.

An initial stylistic description is drafted with GPT-5.1 using the weighted atom evidence alone, with each sentence required to cite its supporting atom IDs. The same model's vision component then inspects the query image as a conservative secondary check: not to rewrite the stylistic account, but to examine the gross scene structure—figures, major objects, setting, and visible condition. This scene information is then compared with the atom prose. Embedding-cosine metadata neighbors supply provisional subject and period context in the same spirit, proposing hypotheses that remain subordinate to atom-cited formal claims. A reasoning-action (ReAct) loop (default three rounds) then stages these sources against one another: atoms continue to organize and ground the formal vocabulary, while vision and neighbor context prune unsupported content and calibrate subject or period framing. This processing yields an atom-cited stylistic description in which dictionary evidence, direct observation, and neighborhood context mutually constrain one another.

For image comparison, the same *Description Agent* is run independently on each painting, freezing the stylistic descriptions for each comparison. The agent first considers whether the works are consistent with the same painter's hand from their shared and differentiating atom structure. Shared atoms and coefficients, for instance, establish common stylistic organization while atoms specific to only one image denote asymmetries. Neither quantity is treated as a calibrated probability of authorship or copying. The output is thus a comparative reading grounded in the *Description Agent*'s evidence.

The Description Agent part of the framework is available as a Google Colab notebook for testing and evaluation:
https://colab.research.google.com/drive/1jSlkOhycELCxL3ITqDXjRHaogm0PlIYD.

**Acknowledgements.** We thank Drs. Lynn Lee and Maryan Ainsworth for giving feedback on a preliminary manuscript draft

**Author contributions**. M.W. and A.H. contributed equally to this manuscript. M.W. and A.H. outlined the architecture and general concepts. M.W. wrote the code and implementation. A.H. selected the case studies. M.W. and A.H. co-wrote the manuscript.

## REFERNCES